\documentclass[11pt]{article}

\usepackage[final]{acl}

\usepackage{times}
\usepackage{latexsym}
\usepackage[T1]{fontenc}
\usepackage[utf8]{inputenc}
\usepackage{microtype}
\usepackage{inconsolata}
\usepackage{graphicx}
\usepackage{booktabs}
\usepackage{xcolor}
\usepackage{pgfplots}
\pgfplotsset{compat=1.18}
\usepgfplotslibrary{groupplots}
\definecolor{domHowto}{HTML}{0072B2}   %
\definecolor{domConv}{HTML}{D55E00}    %
\definecolor{domSocial}{HTML}{009E73}  %
\newcommand{\topl}{$^{\ddagger}$}          %
\newcommand{\subm}{$^{\star}$}             %
\newcommand{\best}[1]{\textbf{#1}}
\definecolor{gainfg}{HTML}{2E7D32}   \definecolor{gainbg}{HTML}{E8F5E9}
\definecolor{dropfg}{HTML}{C43E00}   \definecolor{dropbg}{HTML}{FEF3DE}
\definecolor{samefg}{HTML}{546E7A}   \definecolor{samebg}{HTML}{ECEFF1}
\newcommand{\diffbox}[4]{%
  \tikz[baseline=(char.base)]\node[rounded corners=2pt, fill=#1, text=#2,
    inner xsep=1.6pt, inner ysep=1.1pt, font=\scriptsize\bfseries] (char) {#3#4};%
}
\newcommand{\gain}[1]{\diffbox{gainbg}{gainfg}{$\uparrow$}{#1}}      %
\newcommand{\samedown}[1]{\diffbox{samebg}{samefg}{$\downarrow$}{#1}}
\newcommand{\samenil}[1]{\diffbox{samebg}{samefg}{$\rightarrow$}{#1}}

\title{The University of Melbourne WMT 2026 CreoleMT Submission: \\ A Domain-Balanced Approach to Low-Resource Pacific Creole Machine Translation}

\author{Rapha\"el Merx \hspace{0.5cm} {\bf Nick Thieberger} \hspace{0.5cm} {\bf Ekaterina Vylomova} \\
  The University of Melbourne \\}

\begin{document}
\maketitle

\begin{abstract}
For our submission to the WMT26 Creole Language Translation Shared Task, we focus on machine translation (MT) models for Pacific creoles: Tok Pisin, Bislama, and Solomon Pijin, with particular attention to broad domain performance.
After pre-training on a large collection of domain-imbalanced data, we continue fine-tuning on a diverse mix of domain-balanced data.
We rely on a number of data collection and preparation techniques, including LLM-assisted respelling and alignment, back-translation, and distillation from Gemini for domains originally not present in training data.
Evaluated on Bouquet and a novel test set made of spoken language transcripts, our models beat open model baselines by 3+ chrF++ points in all directions with human-original references.
Looking ahead, we plan to develop human-translated test sets for Solomon Pijin and Bislama, and to distil our best models into much smaller ones that retain broad domain coverage.

\end{abstract}

\section{Introduction}

For this second WMT shared task on Creole machine translation \citep{robinson-etal-2026-findings}, we focus on the three main Pacific creoles: Tok Pisin (tpi, Papua New Guinea), Bislama (bis, Vanuatu), and Solomon Pijin (pis, the Solomon Islands).
They share common characteristics: they are a lingua franca in their respective countries, with widespread use for daily communication \citep{ethnologue-2025}; they are all English-based contact languages from the Melanesian pidgin family \citep{tryon-charpentier-2004}; and for Tok Pisin and Bislama, are official languages of their country \citep{22,23}.

Together, they are spoken by over 10 million people \citep{ethnologue-2025}. However, they have relatively little MT coverage: Google Translate and NLLB only cover Tok Pisin \citep{nllb-2022}, with no consumer-facing model deployment for Bislama or Solomon Pijin. Availability of test sets that cover a wide range of domains (e.g. beyond religion and news) is also limited, with the exception of the recent addition of Tok Pisin to Bouquet \citep{andrews-etal-2025-bouquet}.

As part of this shared task, we focus on training a single model that supports all three languages, with broad domain coverage. To this end, we collate a training set that includes a mix of parallel data from religious sites, but also synthetic data in the educational and informal domains. We rely on Bouquet tpi as our main test set, with the addition of in-house test sets that cover spoken transcripts for all three languages, provided by the Pacific Creole Project \citep{passmore-etal-2025-english}.

We provide the following contributions:

\begin{itemize}
\setlength{\itemsep}{2pt}
\item A single joint model per translation direction covering all three languages which we pre-trained on
  the broad but religion-heavy union of the data that we collected and then continued training on a smaller domain-balanced mix. On Bouquet and the held-out spoken transcripts, it beats every open model baseline by at least 3 chrF++ in all directions with human-original references.
\item A demonstration that distilling a strong closed model is a cheap way to better cover some domains that scraped data lacks, and that the gain saturates early: the first 1.5k
  English$\rightarrow$Tok Pisin pairs bring $+$1.54 chrF++, and the remaining 14.5k add
  $+$1.20.
\item Evaluation using novel held-out spoken transcripts for all
  three languages, showing that performance in this domain is poorly represented by other test sets, and that adding just a few thousand rows of this domain to the training mix provides a substantial lift in it, while not moving performance in other test sets.
\item Three findings that we believe generalise beyond this task: spelling normalisation brings large chrF++ gains, which may not matter as much to end-users as the numbers suggest; domain coverage added in one creole can carry to its
  neighbours; and joint training helps the two lowest-resource languages while leaving the
  highest-resource one unchanged.
\end{itemize}

\section{Related Work}

\paragraph{Creole MT.}
Creole languages have long been under-served by NLP, despite communities' needs that often differ from those assumed by work that simply translates existing English resources \citep{lent-etal-2022-creole}. Recent efforts have started to close this gap: CreoleVal assembles multitask benchmarks spanning 28 creoles \citep{lent-etal-2024-creoleval}, and Kreyòl-MT gathers the largest parallel corpus to date across 41 creoles, finding that a genre-diverse model can outperform a genre-specific one \citep{robinson-etal-2024-kreyol}. The first WMT shared task on creole MT \citep{robinson-etal-2025-findings} brought these threads into a shared evaluation, of which the present task is the second edition. Past findings include: transfer from English is weaker than the surface similarity between the languages suggests \citep{lent-etal-2022-ancestor}; and the available data is dominated by religious text, which transfers poorly to everyday language, while even a few hundred in-domain sentences noticeably lift translation quality \citep{rowe-etal-2025-limitations}. This applies to the Pacific creoles we target, with Tok Pisin the only one to have a broad-domain test set.

\paragraph{Cross-domain low-resource MT.}
Neural MT quality degrades sharply outside its training domain, at times producing fluent but inadequate output \citep{koehn-knowles-2017-six}. The problem is acute for low-resource languages, where the little parallel data that exists is concentrated in a few domains such as religion and news, spurring efforts to build corpora in other domains \citep{merx-etal-2025-openwho}. Domain adaptation for NMT \citep{chu-wang-2018-survey} addresses this with data-centric methods, such as selecting or synthesising in-domain data, and model-centric methods, such as continued fine-tuning. A common recipe first trains on all available, largely out-of-domain data and then fine-tunes on in-domain and back-translated data \citep{imankulova-etal-2019-exploiting}. Back-translating target-side monolingual text \citep{sennrich-etal-2016-improving} and transfer from massively multilingual models \citep{nllb-2022} are the standard levers for widening coverage. We combine these: we start from a massively multilingual model, train it on all collected data, and then continue fine-tuning on a domain-rebalanced mix that adds back-translated news and synthetic educational and informal text to counter the religious skew.

\section{Data}

\subsection{Collection}

\subsubsection{Test sets}
We evaluate on three held-out sets, chosen to reach beyond the religious and news domains
that dominate the available creole corpora.

\paragraph{Bouquet} (854 sentences per language)~\citep{andrews-etal-2025-bouquet} is a broad-domain benchmark spanning
everyday speech, how-to, narration, social-media posts and more. It covers Tok Pisin, but not the other two languages we target, so we build silver references for them using an LLM
(Section~\ref{sec:prep}).

\paragraph{FLORES-200} (2,009 sentences, Tok Pisin only)~\citep{nllb-2022} is the standard multilingual
benchmark: general encyclopaedic and news prose with gold human translations, used as a second
independent Tok Pisin reference point.

\paragraph{Spoken transcripts} (300 sentences per language)~\citep{passmore-etal-2025-english} are transcribed and translated
speech from the Pacific Creole Project, a spoken register not found in other sets and
originally produced creole-to-English. We hold out 300 rows per language for testing and fold the
rest into training (Tok Pisin 10.0k, Bislama 4.5k, Solomon Pijin 1.6k).

\paragraph{Decontamination.} Some of the CreoleMT shared task evaluation set overlaps public text that was in our training data.
Before filtering, most English eval sentences already appeared in our data (75\% for Tok Pisin,
95\% for Bislama, 99\% for Solomon Pijin) as did 66\% of the Tok Pisin creole side (the Solomon
Pijin and Bislama creole sides far less, 3\% and 8\%). We drop from every training corpus any pair
that shares a sentence with an eval source, on either side, after aggressive normalisation;
afterwards no eval sentence remains in our training files.

\subsubsection{Training data}
Table~\ref{tab:data} gives the per-source counts. The data falls into three kinds.

\paragraph{Scraped parallel.} English$\leftrightarrow$creole pairs, consolidated, sentence-aligned
and deduplicated per language, then decontaminated; this forms the training corpus: 472.7k pairs for Tok
Pisin, 42.5k for Solomon Pijin and 80.8k for Bislama. It is heavily religious in domain, with JW300 and the Bible, with,
 for Bislama, 35.1k back-translated Vanuatu news. Verse-level scripture is joined on the canonical
verse id, so those pairs are exact and need no alignment.

\paragraph{Monolingual, back-translated.} Creole-only text machine-translated back into English
(Gemma-4-31B or Gemini) to add colloquial and news register: Tok Pisin news (34.0k sentences) and
web text (FineWeb-2, 18.9k); Vanuatu news and government pages (FineWeb-2, 2.5k documents, 35.2k
pairs) for Bislama; and news for Solomon Pijin (33 documents). Most of the FineWeb-2 data however is not used for our best model training mix, after finding it was detrimental to translation quality (favouring English passthrough).

\paragraph{Synthetic (distilled).} 6.0k English-to-Tok-Pisin pairs generated by a strong closed
model (Gemini 3.5 Flash), sampled evenly from three everyday domains (narration, how-to, social
posts) to supply the colloquial registers the scraped data lacks. Figure~\ref{fig:distill}
measures how far this scales.

\paragraph{Usage.} We train one joint Tok Pisin + Solomon Pijin + Bislama
model per direction. Each first pre-trains on the broad decontaminated union of the data above
(596k pairs: 472.7k Tok Pisin, 42.5k Solomon Pijin, 80.8k Bislama), then continues on a smaller
on-domain mix: scripture and synthetic data for both directions, plus the Tok Pisin high-quality mix
(forward) or held-out spoken transcripts (reverse). Tables~\ref{tab:main} and~\ref{tab:mainpb} give
the full recipe and per-stage scores.

\begin{table}[t]
\centering\small
\setlength{\tabcolsep}{3pt}
\begin{tabular}{lrrr}
\toprule
Source & tpi & pis & bis \\
\midrule
\multicolumn{4}{@{}l}{\emph{Pre-train}} \\
JW300 (OPUS)                        & 371{,}822 & --       & --       \\
Web scrape                       & 72{,}319  & 13{,}809 & 17{,}142 \\
NWT Bible                           & 28{,}526  & 28{,}562 & 28{,}568 \\
News, back-transl.                  & --        & 147      & 35{,}107 \\
\cmidrule(r){1-4}
\textbf{Total}                      & \textbf{472{,}667} & \textbf{42{,}518} & \textbf{80{,}817} \\
\midrule
\multicolumn{4}{@{}l}{\emph{Continued fine-tune (eng$\to$creole)}} \\
NWT Bible                          & 28{,}526 & 28{,}564 & 28{,}568 \\
Book of Mormon                     & 5{,}924  & --       & 5{,}934  \\
Church study library               & 26{,}469 & --       & 27{,}614 \\
jw.org articles              & 47{,}719 & --       & --       \\
\textit{Watchtower} magazine & 39{,}210 & --       & --       \\
News, back-transl.           & 31{,}388 & --       & --       \\
tpi example sentences & 3{,}373  & --       & --       \\
Scraped parallel (re-used)         & --       & 41{,}136 & 52{,}221 \\
Gemini distill.\ (eng$\to$tpi)     & 5{,}991  & --       & --       \\
\cmidrule(r){1-4}
\textbf{Total}                     & \textbf{188{,}600} & \textbf{69{,}700} & \textbf{114{,}337} \\
\midrule
\multicolumn{4}{@{}l}{\emph{Continued fine-tune (creole$\to$eng)}} \\
NWT Bible                          & 28{,}526 & 28{,}564 & 28{,}568 \\
Book of Mormon                     & 5{,}924  & --       & 5{,}934  \\
Church study library               & 26{,}469 & --       & 27{,}614 \\
Spoken transcripts                 & 10{,}035 & 1{,}594  & 4{,}447  \\
Gemini distill.\ (eng$\to$tpi)     & 5{,}991  & --       & --       \\
\cmidrule(r){1-4}
\textbf{Total}                     & \textbf{76{,}945} & \textbf{30{,}158} & \textbf{66{,}563} \\
\bottomrule
\end{tabular}
\caption{Training data mix, in number of sentences. Under eng$\to$creole, the jw.org,
\textit{Watchtower}, back-translated news and example-sentence rows form the Tok Pisin high-quality mix.}
\label{tab:data}
\end{table}

\subsection{Data preparation}
\label{sec:prep}
Three cleaning steps, each applied where a language needed it and each measured as a single-factor
ablation (Table~\ref{tab:abl}).

\paragraph{Loanword respelling (Solomon Pijin).} The scraped Pijin data, almost all from jw.org,
spells English loanwords the English way (``first'', ``change''), whereas everyday and test Pijin use
local spellings (``fas'', ``senis''). We use Gemma-4-31B, guided by a Pijin dictionary, to respell the
training targets into the local convention: spelling only, same words and meaning. This was the
largest single lever we found ($+$7.7 chrF++).

\paragraph{Alignment.} Extract parallel sentences from parallel documents with an LLM aligner (Gemini).
Separately, we pass the Tok Pisin corpus through a Gemma-4-31B judge (using vLLM) that scores each pair and drops the clearly-misaligned ones (490.8k judged, 463.8k
kept). Verse-level Bible and Book of Mormon text is joined using verse id.

\paragraph{Silver test sets (Solomon Pijin, Bislama).} Neither language has a Bouquet set, so we
generate silver references with Gemini from the English source, the Tok Pisin reference, bilingual
dictionary entries (SIL Webonary, 3.0k for Pijin; 4.3k for Bislama) and 20 curated example sentences
per language. Because the reverse-direction source is then model-generated, we mark those cells in our
results tables and do not read them as gold.

\section{Models}

\begin{table*}[t]
\centering
\footnotesize
\setlength{\tabcolsep}{4.5pt}
\begin{tabular}{l ccc ccc}
\toprule
 & \multicolumn{3}{c}{English $\rightarrow$ Tok Pisin} & \multicolumn{3}{c}{Tok Pisin $\rightarrow$ English} \\
\cmidrule(lr){2-4}\cmidrule(lr){5-7}
 & Bouquet & FLORES & Transcript & Bouquet & FLORES & Transcript \\
\midrule
NLLB-3.3B, base                       & 46.7 & 43.1 & 37.8 & 39.2 & 52.4 & 28.7 \\
Krey\`ol-MT                           & 45.6 & 46.7 & 36.3 & 37.7 & 50.9 & 29.7 \\
CreoleM2M                             & 40.8 & 37.9 & 32.8 & 32.2 & 36.1 & 26.1 \\
Gemini 3.5 Flash\topl                 & \best{51.9} & \best{50.2} & 41.0 & \best{50.1} & 52.2 & 50.0 \\
Gemma-4-31B, zero-shot                & 45.4 & 45.5 & 37.1 & 46.1 & 50.7 & 46.5 \\
\cmidrule(lr){1-7}
NLLB, pre-train (all)                 & 46.9 & 43.0 & 38.0 & 45.4 & 53.3 & 36.8 \\
\hspace{0.5em}$+$ continued FT        & 50.0\,\gain{3.1} & 45.7\,\gain{2.7} & 40.8\,\gain{2.8} & 47.6\,\gain{2.2} & \best{55.5}\,\gain{2.2} & 36.1\,\samedown{0.7} \\
\hspace{1.0em}$+$ spoken transcr.     & 49.9\,\samedown{0.1} & 45.5\,\samedown{0.2} & \best{48.7}\,\gain{7.9} & 47.5\,\samedown{0.1} & \best{55.1}\,\samedown{0.4} & \best{53.9}\,\gain{17.8} \\
\hspace{1.5em}$+$ Gemini distill.\subm & \best{51.5}\,\gain{1.6} & 47.6\,\gain{2.1} & \best{47.7}\,\samedown{1.0} & \best{49.2}\,\gain{1.7} & \best{54.6}\,\samedown{0.5} & \best{53.9}\,\samenil{0.0} \\
\addlinespace
Gemma-4-E4B, LoRA pre-train           & 42.1 & 38.4 & 34.1 & --   & --   & --   \\
\quad + cont FT + transcr.\ + distill. & 47.7 & 42.4 & 44.9 & --   & --   & --   \\
Gemma-4-26B, LoRA pre-train           & 42.1 & 38.3 & 34.1 & --   & --   & --   \\
\quad + cont FT + transcr.\ + distill. & 47.4 & 42.4 & 39.9 & --   & --   & --   \\
\bottomrule
\end{tabular}
\caption{
Tok Pisin results, chrF++ on Bouquet (854 human-translated rows), FLORES-200 (2,009 rows) and held-out spoken transcripts (300 rows).
\best{Bold} marks the best score in each column and every score within 1 chrF++ of it (our noise floor); the * row is the system we submit.
Pre-training on all data barely moves the forward direction, but continued fine-tuning and Gemini distillation bring us level with Gemini 3.5 Flash, the strongest baseline, everywhere but FLORES.
Table~\ref{tab:mainpb} reports the same systems and stages for Solomon Pijin and Bislama.
}
\label{tab:main}
\end{table*}

\paragraph{Baselines.} We compare against open multilingual systems that already cover at least one
of our three languages: NLLB-200-3.3B (tpi only, \citealp{nllb-2022}), MADLAD-400-3B (bis/pis only, \citealp{kudugunta-2023-madlad}), Krey\`ol-MT (tpi only, \citealp{robinson-etal-2024-kreyol})
and CreoleM2M (all 3, \citealp{lent-etal-2024-creoleval}). We also report Gemma-4-31B zero-shot and, as a closed frontier model, Gemini
3.5 Flash. Among the systems small enough to fine-tune on a single GPU, NLLB-3.3B is the strongest
base on Tok Pisin, beating others on Bouquet and on the spoken transcripts in both directions (except Tok Pisin$\rightarrow$English transcripts, where
it sits 1 chrF++ below Krey\`ol-MT); Gemma-4-31B is competitive zero-shot but too large to fine-tune
on one GPU. We therefore select NLLB-3.3B for fine-tuning.

\paragraph{Joint multilingual fine-tuning.} We add Solomon Pijin and Bislama target tokens to NLLB
(initialised from the Tok Pisin token) and fine-tune one joint model per direction on all three languages, using the clean data from Section~\ref{sec:prep}.
We also add new tokens for punctuation missing in the NLLB tokenizer (e.g. curly quotes, en/em-dashes).
Training the related languages together beats per-language fine-tunes on the two minority languages (Table~\ref{tab:abl}).

\paragraph{Continued fine-tuning on diverse mix.} Following the joint pre-training above, we continue with fine-tuning on a smaller but more domain-diverse mix. Tables~\ref{tab:main} and~\ref{tab:mainpb} report the per-stage staircase. %

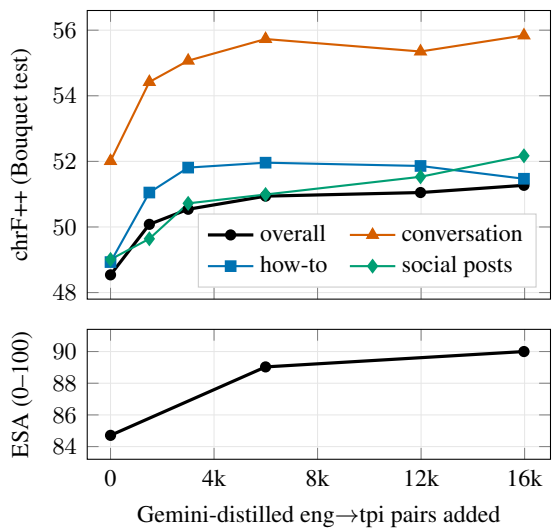
\begin{figure}[t]
\centering
\begin{tikzpicture}
\begin{groupplot}[
  group style={group size=1 by 2, vertical sep=4mm, x descriptions at=edge bottom},
  width=\columnwidth,
  xmin=-900, xmax=17000,
  xtick={0,4000,8000,12000,16000},
  xticklabels={0,4k,8k,12k,16k},
  scaled x ticks=false,
  grid=major, grid style={line width=.2pt, draw=gray!20},
  tick label style={font=\small}, label style={font=\small},
]
\nextgroupplot[
  height=5.4cm, ymin=47.8, ymax=56.6, ytick={48,50,52,54,56},
  ylabel={chrF++ (Bouquet test)},
  legend style={font=\footnotesize, at={(0.97,0.04)}, anchor=south east,
                draw=gray!40, legend columns=2, /tikz/every even column/.append style={column sep=6pt}},
  legend cell align=left,
]
\addplot[very thick, mark=*, mark size=1.5pt, color=black]
  coordinates {(0,48.54) (1500,50.08) (3000,50.54) (5991,50.94) (11978,51.05) (15975,51.27)};
\addlegendentry{overall}
\addplot[thick, mark=triangle*, mark size=2.2pt, color=domConv]
  coordinates {(0,52.01) (1500,54.42) (3000,55.07) (5991,55.73) (11978,55.35) (15975,55.84)};
\addlegendentry{conversation}
\addplot[thick, mark=square*, mark size=1.8pt, color=domHowto]
  coordinates {(0,48.93) (1500,51.05) (3000,51.81) (5991,51.96) (11978,51.86) (15975,51.47)};
\addlegendentry{how-to}
\addplot[thick, mark=diamond*, mark size=2.2pt, color=domSocial]
  coordinates {(0,49.01) (1500,49.64) (3000,50.72) (5991,50.99) (11978,51.53) (15975,52.17)};
\addlegendentry{social posts}
\nextgroupplot[
  height=3.3cm, ymin=83.2, ymax=91.4, ytick={84,86,88,90},
  xlabel={Gemini-distilled eng$\rightarrow$tpi pairs added},
  ylabel={ESA (0--100)}, ylabel style={align=center},
]
\addplot[very thick, mark=*, mark size=1.5pt, color=black]
  coordinates {(0,84.71) (5991,89.03) (15975,90.00)};
\end{groupplot}
\end{tikzpicture}
\caption{Translation quality gains from adding Gemini distillation in ChrF++ points (top) / ESA through LLM-judge (bottom).
Gains are measurable with only 6k sentences (2k each of TweetEval / WikiLingua / ROCStories), then plateau.
}
\label{fig:distill}
\end{figure}

\paragraph{Limited Gemini distillation.} Because Gemini 3.5 Flash performs well on tpi zero-shot,
and because we find very limited domain diversity in scraped data, we distill Gemini 3.5 Flash on
a couple thousand sentences for each of social media text (TweetEval, \citealp{barbieri-etal-2020-tweeteval}), how-to articles (WikiLingua, \citealp{ladhak-etal-2020-wikilingua}) and literature (ROCStories, \citealp{mostafazadeh-etal-2016-corpus}).
For each, we split into sentences, and run Gemini forward from English to Tok Pisin.
We report on the benefit of this distillation in Figure~\ref{fig:distill}.

\begin{table}[t]
\centering
\footnotesize
\setlength{\tabcolsep}{2pt}
\begin{tabular}{l cc cc}
\toprule
 & \multicolumn{2}{c}{eng $\rightarrow$ creole} & \multicolumn{2}{c}{creole $\rightarrow$ eng} \\
\cmidrule(lr){2-3}\cmidrule(lr){4-5}
 & Bouq. & Transc. & Bouq. & Transc. \\
\midrule
\multicolumn{5}{@{}l}{\textbf{Solomon Pijin (pis)}} \\
NLLB-3.3B, base                & 23.9 & 19.3 & 40.8 & 28.3 \\
MADLAD-400-3B                  & 21.4 & 19.2 & 40.0 & 31.6 \\
CreoleM2M                      & 39.2 & 29.5 & 41.5 & 31.1 \\
Gemma-4-31B, zero-shot         & 38.3 & 29.2 & \best{62.3} & 51.4 \\
\cmidrule(lr){1-5}
NLLB, pre-train (all)          & \best{52.8} & 36.6 & \best{62.5} & 45.8 \\
\hspace{0.5em}$+$ continued FT  & \best{52.8} & 37.5 & 61.7 & 43.1 \\
\hspace{1.0em}$+$ spoken transcr. & \best{52.5} & \best{44.2} & 61.8 & \best{55.0} \\
\hspace{1.5em}$+$ Gemini distill.\subm & \best{53.1} & \best{44.0} & \best{63.0} & \best{55.6} \\
\addlinespace
Gemma-4-E4B, LoRA pre-train    & 45.1 & 33.2 & --   & --   \\
\quad + cont FT + transcr.\ + distill. & 46.3 & 37.9 & --   & --   \\
Gemma-4-26B, LoRA pre-train    & 45.0 & 33.1 & --   & --   \\
\quad + cont FT + transcr.\ + distill. & 46.1 & 33.9 & --   & --   \\
\midrule
\multicolumn{5}{@{}l}{\textbf{Bislama (bis)}} \\
NLLB-3.3B, base                & 31.7 & 25.3 & 38.3 & 23.6 \\
MADLAD-400-3B                  & 25.7 & 20.0 & 40.1 & 23.3 \\
CreoleM2M                      & 44.5 & 35.7 & 40.5 & 28.4 \\
Gemma-4-31B, zero-shot         & 51.0 & 40.3 & 57.0 & 46.3 \\
\cmidrule(lr){1-5}
NLLB, pre-train (all)          & 52.9 & 44.3 & 59.0 & 45.0 \\
\hspace{0.5em}$+$ continued FT  & 56.4 & 45.1 & 59.3 & 40.3 \\
\hspace{1.0em}$+$ spoken transcr. & 56.0 & \best{57.8} & 59.3 & \best{54.0} \\
\hspace{1.5em}$+$ Gemini distill.\subm & \best{57.8} & \best{57.5} & \best{60.5} & \best{53.8} \\
\addlinespace
Gemma-4-E4B, LoRA pre-train    & 48.6 & 41.3 & --   & --   \\
\quad + cont FT + transcr.\ + distill. & 53.0 & 51.4 & --   & --   \\
Gemma-4-26B, LoRA pre-train    & 47.8 & 39.7 & --   & --   \\
\quad + cont FT + transcr.\ + distill. & 52.5 & 45.0 & --   & --   \\
\bottomrule
\end{tabular}
\caption{
Solomon Pijin and Bislama results, chrF++; same systems as Table~\ref{tab:main}. Note Bouquet references are Gemini-generated from the Tok Pisin reference (+ lexicon and example sentences, Section~\ref{sec:prep}). Spoken transcripts (300 rows/lang) are human-original in both directions.
Unlike Tok Pisin, these two languages gain most from the broad pre-train, which already beats every baseline.
}
\label{tab:mainpb}
\end{table}

\paragraph{Ablations.}
Table~\ref{tab:abl} isolates each part of our recipe, varying one factor and holding the rest fixed. Gains are measured on Bouquet chrF++ for the en-xx  direction. We also ran a number of data ablations, and ended up using only a limited subset of the data available to us, guided by evaluation on Bouquet chrF++. Data ablations however are mostly not reported in this table.

\begin{table}[t]
\centering
\small
\setlength{\tabcolsep}{5pt}
\begin{tabular}{llcc}
\toprule
Ablation & Lang & w/o\,$\to$\,w/ & $\Delta$ \\
\midrule
Loanword respelling         & pis & 45.8\,$\to$\,53.4 & $+$7.7 \\
Continued FT                & tpi & 46.9\,$\to$\,50.0 & $+$3.1 \\
Gemini distil. (6k pairs) & tpi & 48.5\,$\to$\,50.9 & $+$2.4 \\
Joint training              & bis & 50.7\,$\to$\,52.5 & $+$1.8 \\
                            & pis & 44.4\,$\to$\,45.8 & $+$1.4 \\
                            & tpi & 45.6\,$\to$\,46.3 & $+$0.7 \\
Alignment filtering         & tpi & 46.0\,$\to$\,46.3 & $+$0.3 \\
\bottomrule
\end{tabular}
\caption{Ablations, sorted by effect size; each varies one factor and holds the
rest fixed. Scores are Bouquet en-xx chrF++.
Respelling and alignment filtering use Gemma-4-31B for cleaning.
Continued FT is the two-stage pre-train$\rightarrow$continued-FT recipe from Table~\ref{tab:main}.
Gemini distillation adds 6k synthetic eng$\rightarrow$tpi pairs (the 0 and 6k points of Figure~\ref{fig:distill}).
The joint-training rows compare each language's single-language NLLB (w/o)
against the joint model (w/) on raw-spelling data, so a positive $\Delta$ means joint training helps
that language.}
\label{tab:abl}
\end{table}

\section{Results}

\paragraph{Pretrain.}
NLLB-3.3B (base) sits 5.2 chrF++ behind Gemini 3.5 Flash on Tok Pisin Bouquet (46.7
vs 51.9). The first stage (broad pre-train on all data) adds almost nothing on Tok Pisin in the forward
direction (46.7 to 46.9), but does in the reverse ($+$6.2, 39.2 to 45.4).
For the other two languages, the pretrain is enough to
beat every baseline we compare against: 52.8 on Pijin Bouquet, 13.6 above CreoleM2M, and 52.9 on
Bislama, 8.4 above it. These Bouquet references are silver (Section~\ref{sec:prep}) and should be
read with care, but the spoken transcripts are human-original, and they show gains of
the same size over CreoleM2M ($+$7.1 for Pijin, $+$8.6 for Bislama), so we read the improvement as
real.

\paragraph{Continued fine-tuning.}
The three continued fine-tuning stages (domain-balanced data, spoken transcripts, and Gemini distillation) close the gap with Gemini-3.5-Flash (the best model we measured for Tok Pisin):
our shipped model scores 51.5 against Gemini's 51.9 on Bouquet, and pulls ahead on the spoken domain, 47.7 against 41.0 on transcripts.
FLORES is the exception, where we stay 2.6 points behind (47.6 against 50.2), but FLORES' age and widespread use may mean contamination in recent models \citep{tan-etal-2026-flores}.
In the reverse direction the shipped model is 0.9 below
Gemini on Bouquet (49.2 against 50.1) and 3.9 above it on transcripts (53.9 against 50.0).

\paragraph{Fine-tuning LLMs: promising but expensive.}
We also try LoRA fine-tuning of Gemma-4 using the same two-stage recipe, at two sizes (E4B and 26B). Both
land below the NLLB fine-tunes: 47.7 and 47.4 on Tok Pisin Bouquet against 51.5 for our
shipped model, with the same ordering on Pijin (46.3 and 46.1 against 53.1) and Bislama (53.0 and
52.5 against 57.8). Going from E4B to 26B does not improve the scores, which suggests the limit is the
LoRA recipe, not model capacity.
Gemma-4-31B, on the other hand, is competitive with no fine-tuning at all: zero-shot it reaches 45.4 on Tok Pisin, just below base
NLLB, 51.0 on Bislama, above every open baseline we test. Fine-tuning a model of
that size therefore looks like a promising direction, but it does not fit our single-GPU budget, so
we leave it to future work.

\paragraph{Domain diversity.}
Breaking Tok Pisin Bouquet down by domain shows where the base model falls short and where
fine-tuning helps most. Out of the box, NLLB trails Gemini most on the informal, conversational
registers that our scraped corpora (religion and news) barely contain: it is 7.4 chrF++ behind on
conversation (48.4 against 55.9), 6.5 on user comments, 5.3 on narration and 5.0 on social posts,
against only 2.5 on reflective prose. Our fine-tune recovers most of this gap: it now beats Gemini on
how-to (52.9, 2.4 above) and draws level on conversation (56.3 against 55.9) and social posts (51.5
against 52.2). The registers where we still trail are the ones our training data covers least, and
only by 1 to 3 points: comments (47.0 against 49.8) and narration (52.1 against 53.6). The remaining
half-point deficit to Gemini on Bouquet is thus concentrated in a few under-covered registers rather
than spread evenly.

\subsection{Discussion}

\paragraph{Distillation is cheap and beneficial.}
  Figure~\ref{fig:distill} scales the Gemini distillation set from 0 to 16k pairs, holding the
  122.4k human tpi mix and the recipe fixed. The curve rises steeply and then flattens: the first
  1.5k pairs are worth $+$1.54 chrF++, more than half of the $+$2.73 that 16k pairs buy in total,
  and the remaining 14.5k pairs add only $+$1.20 between them. No single increment past the first
  is individually resolvable (1.5k$\rightarrow$3k, 3k$\rightarrow$6k, 6k$\rightarrow$12k and
  12k$\rightarrow$16k all have 95\% intervals spanning zero), yet the accumulated
  1.5k$\rightarrow$16k gap is significant, so the plateau slopes gently upward rather than being
  flat. The takeaway is that distillation from a strong closed model is cheap to exploit
  and quick to exhaust: a few thousand well-chosen source sentences capture most of the available
  gain, and buying an order of magnitude more yields roughly one further chrF++ point.
  Beside ChrF++, we also score with a reference-based LLM judge (GPT-5.6-luna, ESA-style 0--100 with error spans). 
  The LLM-judge also measures translation quality gains: $+$4.3 ESA from 0 to 6k with a 95\% interval clear of
  zero, and a further $+$1.0 to 16k that is not resolvable.

\paragraph{Adding transcripts reduces over-translation of spoken text.} On the (in-domain)
  spoken transcript test set, the reverse model trained without transcripts tends to render
  terse spoken text as more formal, idiomatic English, whereas the transcript-augmented model
  mirrors the literal register of the references. For example, for
  \emph{stap klostu wantaim mi olgeta taim} (ref.\ ``stay close to me at all times''), the
  no-transcript model outputs the idiomatic ``always by my side'', while the +transcripts model
  gives the literal ``stay close to me all the time''.

\paragraph{The outsize weight of orthography.} Spelling consistency between training data and
  reference drives chrF++ to an extent that may not reflect end-user priorities. Respelling pis to the reference
  convention gained \textbf{$+$7.7} (45.8 to 53.4), while the two languages left unrespelled in the
  same run barely moved (bis $-$0.2, tpi $+$0.3). Adding more religious-register data then
  helped bis, whose orthography matches ($+$2.1), but hurt pis ($-$3.6), whose text uses a different yet valid
  convention (\texttt{bilong} rather than \texttt{blong}): a 5.7 chrF change from spelling alone,
  against $+$2.0 for our best modelling change. As \texttt{bilong} is perfectly comprehensible Pijin,
  and the dominant spelling in the Pijin religious publications we scrape, we question whether chrF++ measures quality or mere
  conformity to the reference's spelling, and whether optimising for it trades away subtler gains in
  terminology or register.

\paragraph{Domain coverage added in one creole may carry to its neighbours.} The final stage adds 
  a tpi-only Gemini distillation set. Beyond the $+$1.6 it adds on Tok Pisin
  Bouquet (49.9 to 51.5), it also improves the silver \textbf{bis} Bouquet set by \textbf{$+$1.8}
  (56.0 to 57.8) and nudges \textbf{pis} up $+$0.6 (52.5 to 53.1), although neither receives new
  training data at that stage. A plausible explanation is that broadening domain coverage in one
  language generalises to the other two sharing the joint model. Two
  caveats here: the added tpi pairs also enlarge the joint
  mix, so every language sees marginally more optimisation steps, and the silver bis references are also
  Gemini-generated, as is the distillation data, so some of the gain may be convergence toward the
  reference author's style rather than improved translation.

\paragraph{Joint training mainly helps the two lowest-resource languages.} Fine-tuning
  the three creoles together rather than separately (holding data and recipe fixed) gains
  \textbf{$+$1.8} for bis (50.7 to 52.5) and \textbf{$+$1.4} for pis (44.4 to 45.8), but only
  \textbf{$+$0.7} for tpi (45.6 to 46.3), inside our $1$ chrF++ noise floor
  (Table~\ref{tab:abl}). We hold far more tpi data, and across more domains, than pis or bis
  (463.8k cleaned pairs against 83k for bis and 44.7k for pis), so joint training lets the two
  data-poor languages draw on the larger and more varied tpi corpus through the shared model,
  while tpi is already well covered on its own and has little left to gain. This is the
  cross-creole transfer of the previous finding seen from the data side, and it is why we train one
  joint model rather than three: the low-resource languages benefit and the high-resource one is
  not harmed.

\section{Conclusion}

In this work, we develop a first model that focuses on Pacific creoles (Tok Pisin, Solomon Pijin, Bislama), targeting good performance across domains. We show that pre-training on an unbalanced mix of data, then fine-tuning on a more balanced and diverse mix can bring substantial translation quality gains across domains, and that these cross-domain gains can transfer across languages (e.g. tpi diverse bringing quality gains to pis/bis). We also show the value of forward-distillation from a strong LLM, on domains that are otherwise unrepresented in available training data.

In future work, we plan to develop more reliable test sets for Solomon Pijin and Bislama, as we feel like Gemini-generated test sets, even though they are guided by Tok Pisin ground truth and a lexicon, are unsatisfactory. We also plan to develop a very small (< 50M) model that can translate across 6 directions, and that has good performance across domains.

\bibliography{custom,anthology-1,anthology-2}

\end{document}